\documentclass[10pt,twocolumn,letterpaper]{article}
\usepackage{cvpr} 
\usepackage{graphicx}
\usepackage[accsupp]{axessibility}
\usepackage{import}
\usepackage{amsmath}
\usepackage{amssymb}
\usepackage{booktabs}

\usepackage{adjustbox}
\usepackage{tabularx}

\usepackage{color}
\usepackage[dvipsnames]{xcolor}
\usepackage{enumitem}

\usepackage{booktabs}
\usepackage{multirow}
\usepackage{colortbl}
\definecolor{Gray}{gray}{0.9}
\makeatletter
\@namedef{ver@everyshi.sty}{}
\makeatother
\usepackage{tikz}

\newrobustcmd*{\mytriangle}[1]{\tikz{\filldraw[draw=#1,fill=#1] (0,0) --
(0.2cm,0) -- (0.1cm,0.2cm);}}

\usepackage[pagebackref,breaklinks,colorlinks]{hyperref}

\usepackage[capitalize]{cleveref}
\crefname{section}{Sec.}{Secs.}
\Crefname{section}{Section}{Sections}
\Crefname{table}{Table}{Tables}
\crefname{table}{Tab.}{Tabs.}

\usepackage[dvipsnames]{xcolor}
\usepackage{xcolor-material}
\usepackage{color, colortbl}
\usepackage{soul}

\def\cvprPaperID{7477} 
\def\confName{CVPR}
\def\confYear{2022}

\begin{document}

\title{\vspace{-0.7cm}E$^2$(GO)MOTION: Motion Augmented Event Stream \\ for Egocentric Action Recognition}

\author{
Chiara Plizzari\thanks{The authors equally contributed to this work.  }\textsuperscript{ }\textsuperscript{          ,1}
 \quad
Mirco Planamente\footnotemark[1]\textsuperscript{ }\textsuperscript{          ,1,2} \quad
Gabriele Goletto\textsuperscript{1} \quad
Marco Cannici\textsuperscript{3}
\quad
Emanuele Gusso\textsuperscript{1} \\
\quad
Matteo Matteucci\textsuperscript{3}
\quad
Barbara Caputo\textsuperscript{1,2}
\and \textsuperscript{1} Politecnico di Torino \\
{\tt\small {name.surname}@polito.it}
\and \textsuperscript{2}  CINI Consortium
\and \textsuperscript{3}  Politecnico di Milano \\
{\tt\small {name.surname}@polimi.it}
}

\maketitle

\begin{abstract}

Event cameras are novel bio-inspired sensors, which asynchronously capture pixel-level intensity changes in the form of “events". Due to their sensing mechanism, event cameras have little to no motion blur, a very high temporal resolution and require significantly less power and memory than traditional frame-based cameras.
These characteristics make them a perfect fit to several real-world applications such as egocentric action recognition on wearable devices, where fast camera motion and limited power challenge traditional vision sensors. However, the ever-growing field of event-based vision has, to date, overlooked the potential of event cameras in such applications.
In this paper, we show that event data is a very valuable modality for egocentric action recognition. To do so, we introduce N-EPIC-Kitchens, the first event-based camera extension of the large-scale EPIC-Kitchens dataset. In this context, we propose two strategies: (i) directly processing event-camera data with traditional video-processing architectures (E$^2$(GO)) and (ii) using event-data to distill optical flow information (E$^2$(GO)MO).
On our proposed benchmark, we show that event data provides a comparable performance to RGB and optical flow, yet without any additional flow computation at deploy time, and an improved performance of up to $4\%$ with respect to RGB only information.
The N-EPIC-Kitchens dataset is available at \small{\url{https://github.com/EgocentricVision/N-EPIC-Kitchens}}.


\end{abstract}

\section{Introduction}
\label{sec:intro}


Egocentric vision has introduced a variety of new challenges to the computer vision community, such as human-object interaction~\cite{damen2014you,lu2021understanding}, action anticipation~\cite{abu2018will,furnari2020rolling,girdhar2021anticipative,liu2020forecasting}, action recognition~\cite{kazakos2019epic}, and video summarization~\cite{del2016summarization,lee2012discovering,lee2015predicting}. With the advent of novel large-scale datasets~\cite{damen2018scaling,damen2021rescaling}, new tasks are being proposed, such as wearer's pose estimation~\cite{wen2021seeing} and egocentric videos anonymization~\cite{thapar2021anonymizing}. This trend will grow in the next years thanks to the very recent release of Ego4D~\cite{grauman2021ego4d}, a massive-scale egocentric video dataset offering more than 3,000 hours of daily-life activity videos accompanied by audio, 3D meshes of the environment, eye gaze, stereo, and multi-view videos.


\begin{figure}
    \centering
    \includegraphics[width=0.9\columnwidth]{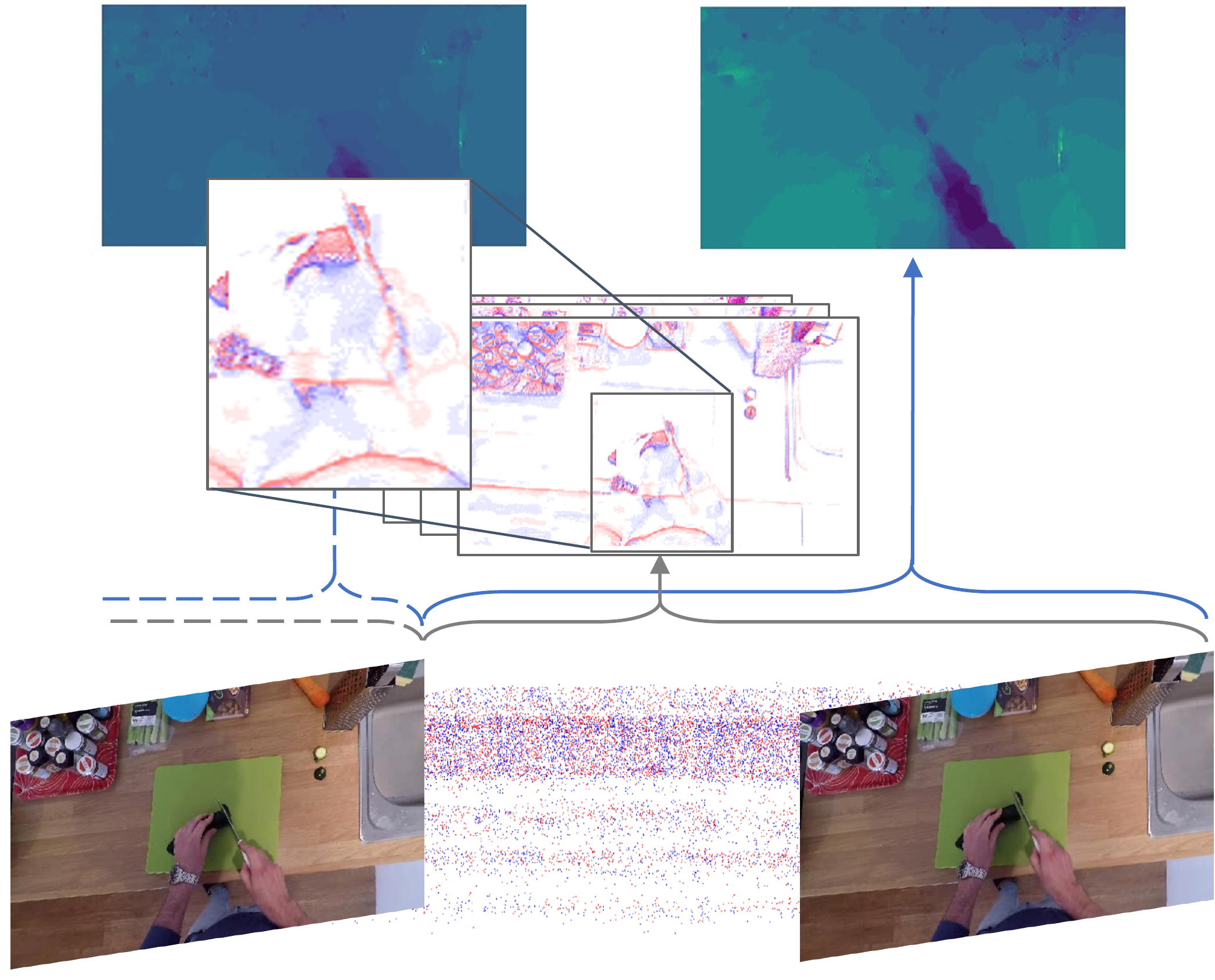}
    \caption{ \textbf{N-EPIC-Kitchens}: the first event-based dataset for egocentric action recognition. From RGB images, we generate a stream of events (bottom). Positive polarity is represented by red events, whereas blue events represent negative polarity. Events focus on motion, similarly to  optical flow (top). With their low latency, high temporal resolution, and low-power consumption, event data are a perfect fit for egocentric action recognition. }
    \vspace{-0.5cm}
    \label{fig:teaser}
\end{figure}

Among all, RGB sensors provide by far the richest source of visual information. However, the performance of RGB-based deep models drastically decrease when the training and test data do not share the same distribution \cite{david2010impossibility}.
This issue, known as environmental bias \cite{munro2020multi, planamente2021domain, song2021spatio, kim2021learning, sahoo2021contrast}, originates from RGB-based networks' tendency to rely on the environment in which activities are recorded, affecting their ability to recognize actions when they are performed in unfamiliar (unseen) surroundings. 
This is mainly caused by appearance-based networks' tendency to primarily focus on background cues and objects texture, which are typically uncorrelated with the action being performed and thus largely varying in different environments. As a result, appearance-free modalities, such as motion, have become the favored choice in current egocentric vision systems, as testified by the results of recent EPIC-Kitchens challenges \cite{ek19report, ek2020, ek2021}. However, the optical flow used in this setting is computed from RGB frames by solving expensive optimization problems (TV-L1 algorithm \cite{zach2007duality}), introducing significant test-time computations~\cite{crasto2019mars}.

Event-based cameras, on the other hand, have been shown to be particularly suitable for online settings \cite{Delbruck2016Neuromorophic,gallego2019event}. Their high pixel bandwidth results in reduced motion blur, and the extremely low latency and low power consumption make these novel sensors particularly good in egocentric scenarios, where fast motion often impacts RGB-based systems negatively. Moreover, as they only convey differential information, event sequences reveal more information about the dynamic of the scene than its appearance, making them a valid alternative to RGB frames when learning to focus on motion.
Still, despite these advantages, no prior research has looked at how to exploit their sensitivity to motion in egocentric vision, where these devices remain unused.

As a first step in this direction,  we propose N-EPIC-Kitchens, a novel dataset that enables, for the first time, the use of event data in this context. It consists in the extension of the large-scale EPIC-Kitchens dataset~\cite{damen2018scaling} under the setup proposed in~\cite{munro2020multi}. The latter is particularly appealing for both the availability of multiple environments (kitchens) and multiple modalities, i.e., RGB, optical flow, and audio. 
These characteristics allow for the analysis of the aforementioned environmental bias as well as the comparison of event data to well-established modalities.
On the proposed N-EPIC-Kitchens, we introduce two approaches to exploit the intrinsic motion characteristics of event data in this context. The first, which we call E$^2$(GO), consists in extending traditional 2D and 3D action recognition architectures with layer variations aimed at exploiting the motion-rich features of event data. The second, E$^2$(GO)MO, extends motion reasoning by distilling motion information from optical flow to event data.  
This is accomplished following a teacher-student approach that allows taking full advantage of expensive offline TV-L1 flow during training only, while avoiding its computation at test time.  We summarize our contributions as follows:
\begin{itemize}
    \item We release N-EPIC-Kitchens, the first event-based egocentric action recognition dataset, which unlocks the possibility to explore event data in this context; 
    \item We benchmark N-EPIC-Kitchens on popular action recognition architectures, showing performance of both event data alone and combined with RGB and optical flow modalities. Moreover, we demonstrate the robustness of event data to environment changes; 
    \item We propose E$^2$(GO) and E$^2$(GO)MO, two event-based approaches tailored at emphasizing motion information captured by event data in egocentric action recognition;
    \item We show that event data can outperform RGB in challenging unseen environments and are competitive with them in known environments, suggesting that using event data is a viable option and more research should be performed in this direction.
    
\end{itemize}

\section{Related Works}
\label{sec:RW}

\paragraph{Event-based Vision.}
Taking advantage of the event-based cameras' inherent ability to perceive changes~\cite{gallego2019event,Delbruck2016Neuromorophic}, researchers have started creating new solutions to tackle traditional computer vision problems exploiting this new way of sensing the world, including optical flow prediction~\cite{events-zhu2019unsupervised,Gehrig3dv2021}, motion segmentation~\cite{parameshwara20210,Zhou21tnnls}, depth estimation~\cite{HidalgoCarrio2020Learning,Gehrig2021Combining}, and many others. While traditional cameras are capable of providing very rich visual information at the tradeoff of slow and often redundant updates, event-based cameras are asynchronous and spatially sparse, and capable of microseconds temporal resolution. Event-based systems range from designs that focus on exploiting and maintaining event-camera sparsity during computation~\cite{bi2020graph,yang2019modeling,sekikawa2019eventnet}, to algorithms that combine events with standard cameras~\cite{hu2020ddd20,planamente2021da4event,cannici2021n,tulyakov2021time,Gehrig2021Combining}, exploiting the complementarity of the two. With the goal of achieving minimum-delay computing, research has also focused on asynchronous designs, either by modifying regular CNNs~\cite{messikommer2020event,cannici2019asynchronous} or by utilizing specific hardware solutions~\cite{furber2012overview,akopyan2015truenorth,davies2018loihi}, often leveraging on bio-inspired computing frameworks~\cite{events-maass1997networks}.
Despite event-based cameras have already been applied to action and gesture recognition tasks~\cite{events-innocenti2020temporal,chen2020dynamic,lungu2019incremental}, previous works have not taken advantage of their complementarity with other visual modalities yet in these contexts, and used these cameras mainly in controlled environments where both the camera and the background are static~\cite{amir2017low,miao2019neuromorphic}. In this paper, instead, we tackle egocentric action recognition with events for the first time and combine them with other modalities. 
\\[2pt] \textbf{Action Recognition.}
The success of 2D CNNs in the context of image recognition~\cite{he2016deep, ioffe2015batch} inspired the first video understanding architectures. Traditional 2D CNNs are often used to process frames individually, eventually fusing optical flow information~\cite{wang2018temporal}, while late fusion mechanisms ranging from average pooling~\cite{wang2016temporal}, multilayer perceptrons \cite{zhou2018temporal}, recurrent aggregation \cite{donahue2015long, li2018videolstm}, and attention~\cite{girdhar2017attentional, sudhakaran2019lsta} are employed to model temporal relations for action understanding.
The use of 3D convolutions has also been proposed as an alternative~\cite{carreira2017quo, tran2015learning}. 
However, despite their ability to learn spatial and temporal relations simultaneously, they often introduce more parameters, requiring pre-training on large-scale video datasets~\cite{carreira2017quo}.
To reduce the model's complexity, other approaches focus on finding more efficient architectures~\cite{tran2019video, sun2015human,feichtenhofer2020x3d, tran2018closer, xie2018rethinking, qiu2017learning}. As an example, a parameter-free channel-wise temporal shift operator has been introduced in the Temporal Shift Module (TSM) network~\cite{lin2019tsm}, resulting in a 2D CNN capable of encoding temporal information. 
Although all these architectures aim at implicitly modeling motion, most of them still mix video frames with the externally estimated optical flow. While this improves the overall performance, it also requires pre-computing the flow, making these approaches impracticable in online settings. In addition, two-stream approaches come at the cost of increased model complexity and number of parameters. To overcome this issue, a line of research proposes approaches that integrate the RGB and optical flow modalities in lighter architectures~\cite{zhao2019dance, lee2018motion, wang2019self}. 
Finally, authors of~\cite{crasto2019mars, planamente2021self} proposed to distill optical flow information to the RGB stream at training time, while avoiding flow computation at test time. 
%
%
%
%
%
\\[2pt] \textbf{First Person Action Recognition.}
The complex nature of egocentric videos raises a variety of challenges, such as ego-motion~\cite{li2015delving}, partially visible or occluded objects, and environmental bias~\cite{munro2020multi, planamente2021domain, song2021spatio, kim2021learning, sahoo2021contrast}, which limit the performance of traditional, third-person, approaches when used in first person action recognition (FPAR)~\cite{damen2018scaling, damen2021rescaling}.
The community's interest has quickly grown~\cite{ek19report, ek2020, ek2021, rodin2021predicting} in recent years, thanks to the possibilities that these data open for the evaluation and understanding of human behavior,
leading to the design of novel architectures~\cite{kazakos2019epic, sudhakaran2019lsta, wang2021interactive, kazakos2021little, furnari2020rolling}.
While the use of optical flow has been the de-facto procedure~\cite{damen2018scaling, damen2021rescaling, ek19report, ek2020, ek2021, grauman2021ego4d} in FPAR, the interest has recently shifted towards more lightweight alternatives, such as gaze~\cite{li2021eye,fathi2012learning, min2021integrating}, audio~\cite{kazakos2019epic,cartas2019seeing, planamente2021domain}, depth~\cite{garcia2018first}, skeleton~\cite{garcia2018first }, and inertial measurements~\cite{grauman2021ego4d}, to enable motion modeling in online settings. These, when combined with traditional modalities, produce encouraging results, but not enough to make them viable alternatives. 
With this work, we show that the intrinsic motion information encoded by event data makes this modality potentially more suitable than RGB.

\section{N-EPIC-Kitchens}
\begin{figure}
    \centering
    \includegraphics[width=\columnwidth]{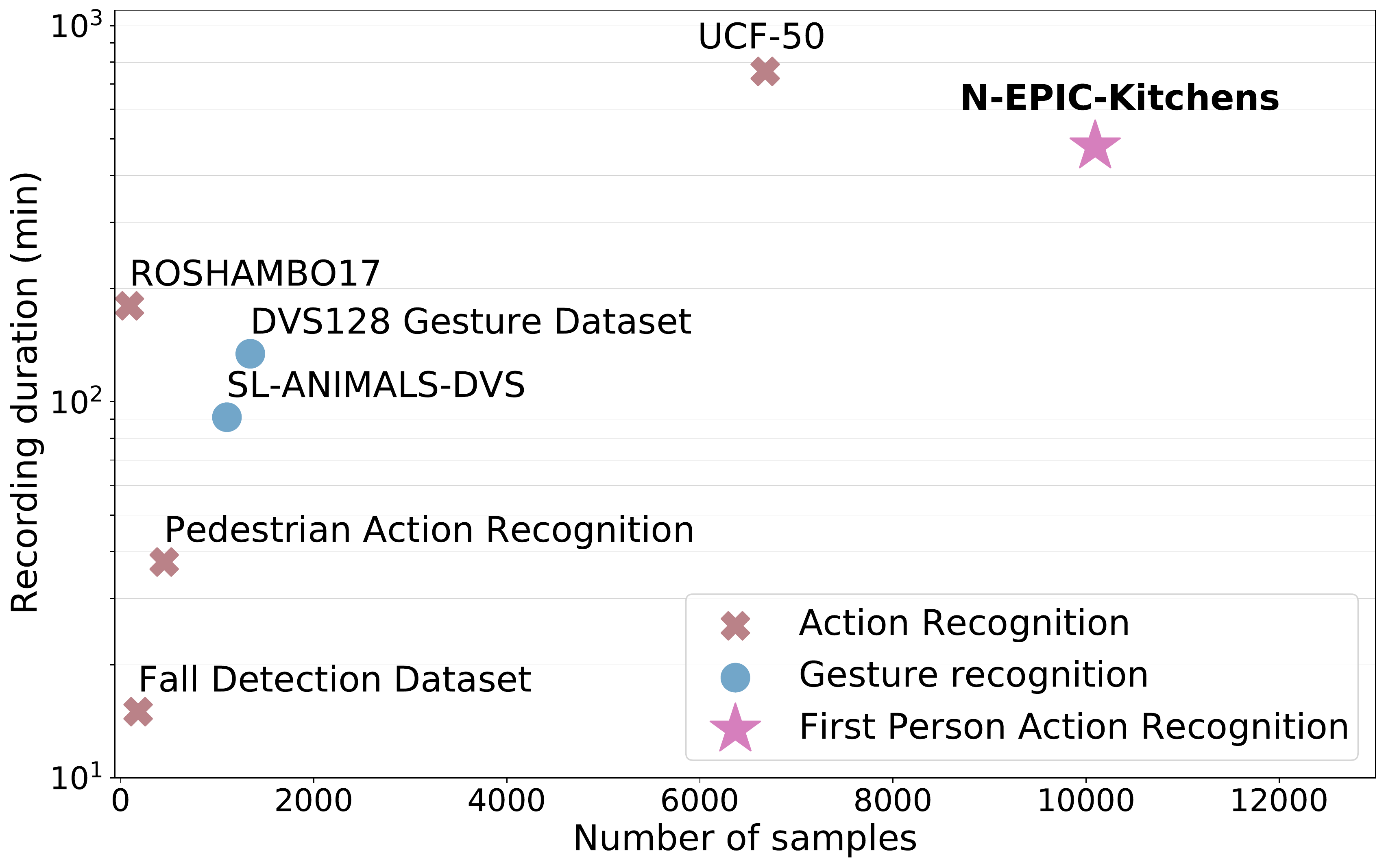}
    \caption{N-EPIC-Kitchens \textit{vs }existing event-based action classification datasets in the literature \cite{roshambo, hu2016dvs, miao2019neuromorphic,amir2017low,vasudevan2020introduction}.}
    \vspace{-0.4cm}
    \label{fig:dataset}
\end{figure}
Thanks to their focus on capturing only variations in the scene, event-based cameras are particularly efficient in 
egocentric scenarios, as they drastically reduce the amount of data to be processed and acquired, avoiding motion blur artifacts and providing fine-grained temporal information. However, 
so far only a limited amount of datasets have been made freely accessible~\cite{hu2016dvs,de2020large,perot2020learning,gehrig2021dsec}. Despite the field is actively working towards increasing their availability, as testified by the recent release of event-based versions of ImageNet~\cite{kim2021n,lin2021esimagenet}, relatively few datasets for human activity recognition are currently available. As reported in Figure \ref{fig:dataset}, most of them focus on action or gesture recognition ~\cite{miao2019neuromorphic,amir2017low,events-innocenti2020temporal,hu2016dvs} in controlled settings, where both the camera and the background are static, and none considers egocentric action recognition, preventing event-based cameras use in this scenario.
To demonstrate the advantages of event-based cameras in egocentric online settings, as well as their complementarity and equivalence to other modalities, we extend the EPIC-Kitchens (EK)~\cite{damen2018scaling} dataset, a large collection of egocentric action videos featuring multiple modalities and different environments. Following the setting of \cite{munro2020multi}, we selected the three largest kitchens from EPIC-Kitchens in number of training action instances, which we refer to as D1, D2 and D3, analysing the performance for the 8 largest action
classes, i.e., ‘put’, ‘take’, ‘open’, ‘close’, ‘wash’, ‘cut’, ‘mix’ and ‘pour’.

In the following, we first introduce the operating principles of DVS cameras. Then, we outline the approach used to generate N-EPIC-Kitchens and emphasize its benefits.





\subsection{Event-Based Vision Data}
Pixels of DVS cameras are independent and respond to changes in the continuous log brightness signal $L(\textbf{u}, t)$, differently from a standard RGB camera.  An event is a tuple $e_{k} = (x_{k}, y_{k}, t_{k}, p_{k})$ specifying the time $t_{k}$, the location $(x_{k},y_{k})$ and the polarity $p_{k} \in \{-1,1\}$ of the bright change (brightness decrease or decrease). An event is triggered when the magnitude of the log brightness at pixel $u = (x_{k}, y_{k})^{T}$ and time $t_{k}$ has changed by more than a threshold $C$ since the last event at the same pixel, as described in the following equation:
\begin{equation}
 \Delta L(\mathbf{u}, t_{k}) = L(\mathbf{u}, t_{k}) - L(\mathbf{u},t_{k} - \Delta t_{k}) \geqslant  p_{k}C .
 \label{eq:event}
\end{equation}
Therefore, the output of an event camera is a continuous stream of events described as a sequence $\mathcal{E}=\{(x_{k}, y_k, t_k, p_k)| t_k \in \tau \}$, being $\tau$ the time interval. 


\paragraph{N-EPIC-Kitchens generation.}


\begin{figure}
    \centering
    \includegraphics[width=0.8\columnwidth]{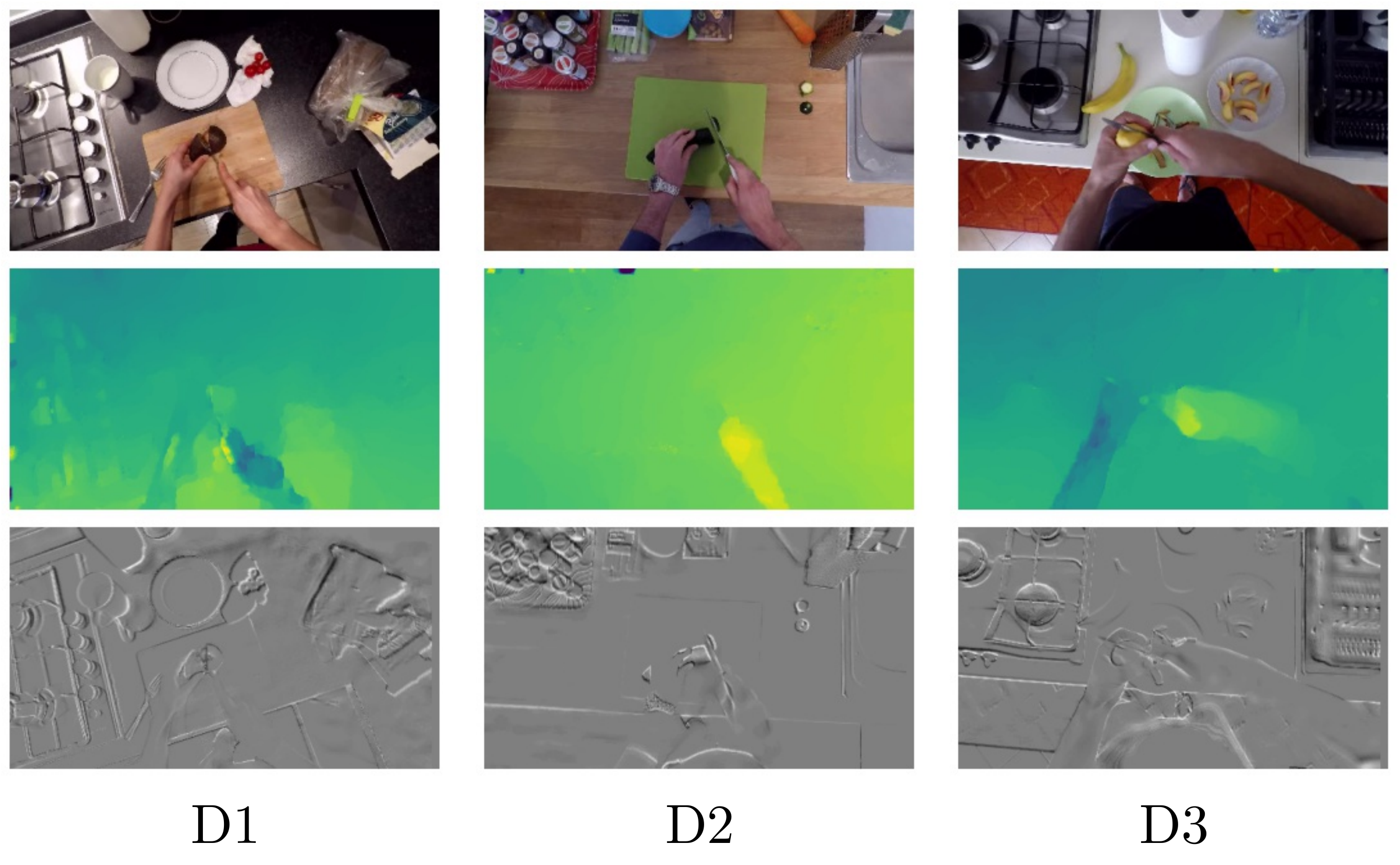}
    \caption{RGB (top), optical flow (middle) and Voxel Grid representation (bottom) from the same action (``cut") on the three different kitchens (D1, D2, D3). }
    \label{fig:kitchens}
\end{figure}

We leverage ESIM~\cite{rebecq2018esim}, a recent event camera simulator, to enhance the EPIC-Kitchens dataset with the event modality. Since videos in EPIC-Kitchens are limited to $60$ frames per second, far lower than the microseconds temporal resolution of an event camera, we first upsample them to a higher fps. To this end, we used Super SloMo~\cite{jiang2018super} for its unique ability to generate frames at any temporal precision, following the adaptive sampling procedure proposed in Vid2E~\cite{gehrig2020video} to extract event streams.
Finally, we use Voxel Grid~\cite{events-zhu2019unsupervised}, a frame-like event encoding technique, to convert sparse and asynchronous events to a tensor representation and enable learning with typical convolutional neural network architectures. 

\section{Challenges of Evaluating Event Data}
\label{ChallengesEventData}

The fundamental problem in assessing event data in first-person action recognition comes from the fact that, unlike other modalities, its use in egocentric vision is completely novel. To set a benchmark in this setting, we evaluate four different aspects of event-based modeling. We start by considering the importance of performance on both seen and unseen test sets, where \textit{seen} indicates performance on the same kitchen on which training is performed, and \textit{unseen} the performance obtained on a different one. We propose to evaluate them altogether in our experiments. While the first provides a good indication of the modality's upper bound performance, the second evaluates the ability of the model to encode domain invariant features and, as a result, the viability of using it in real-world scenarios. Then, as the performance of different modalities may greatly vary depending on the architecture used for processing~\cite{price2019evaluation}, we benchmark events using three of the most accredited architectures in FPAR, namely TSM~\cite{lin2019tsm}, TSN~\cite{wang2018temporal} and I3D~\cite{carreira2017quo}. We leverage a well-established procedure for converting event streams into a frame-like representation that has been shown to efficiently integrate with off-the-shelf CNNs~\cite{stoffregen2020reducing,planamente2021da4event}, and finally propose to encourage modeling of motion features by employing attention at channel level. 
\\[4pt] \textbf{Event Representation.} 
Since event cameras produce sparse encodings of the scene, they must be  converted into intermediate representations before processing. Several representations have been proposed, ranging from bio-inspired~\cite{events-maass1997networks,events-cohen2016thesis,cannici2019asynchronous} to more practical ones. Frame-like representations are by far the most widespread methods as they can be directly used together with off-the-shelf networks. Among available ones~\cite{events-lagorce2016hots,events-sironi2018hats, events-zhu2019unsupervised,cannici2019asynchronous, events-innocenti2020temporal,events-gehrig2019end, events-cannici2020differentiable,events-deng2020amae} we chose Voxel Grid~\cite{events-zhu2019unsupervised} as it proved to be superior in cross-domain settings~\cite{stoffregen2020reducing,planamente2021da4event}.
%
This representation computes a $B$-channel image by discretizing time in $B$ separate intervals:
\begin{equation}
    \mathbf{x}^E(x, y, b) = \sum_{k=1}^N{p_k k_b(b - t_k^*)},
\end{equation}
where $b$ are the channels, $t_k^*$ are the timestamps scaled into $[0, B-1]$, $p_k$ is the polarity and $k_b(a) = max(0,1-|a|)$. 
%
\\[4pt] \textbf{Backbone Architectures.} To assess how event data behaves on different network designs, we examine two popular 2D-CNN approaches, TSM~\cite{lin2019tsm} and TSN~\cite{wang2018temporal} as well as one 3D-CNN, I3D~\cite{carreira2017quo}. The first two rely on a 2D-CNN backbone, but while TSN~\cite{wang2018temporal} can only leverage late fusion for temporal modeling, TSM~\cite{lin2019tsm} exploits \textit{shift modules} to exchange channel information across adjacent frames. In contrast, I3D~\cite{carreira2017quo} is a pure 3D-CNN model, which \textit{inflates} filters and pooling kernels into the temporal dimension. In the literature, there is currently no clear winner, as some modalities may react better with one technique than the other indiscriminately.
%
\\[4pt] \textbf{The Importance of Motion.} 
Environmental biases are typically managed in egocentric vision systems by employing complementary, often appearance-free, modalities. Optical flow is generally the one performing the best in action recognition tasks \cite{damen2021rescaling,damen2018scaling,wang2018temporal}, as (i) it helps focusing on the moving content, i.e., the action being performed, while (ii) preserving the edges of moving objects and (iii) ignoring background information. 
In this paper, we argue that event cameras' sensitivity to moving edges and ability to disregard static information only partially capture the three key features of optical flow listed above. In reality, as a result of the camera movement, these sensors still catch events in the background. This encourages us to learn from flow in order to improve our ability to filter out less discriminative data.

\section{Learning from Motion}

\begin{figure}
    \centering
    \includegraphics[width=\columnwidth]{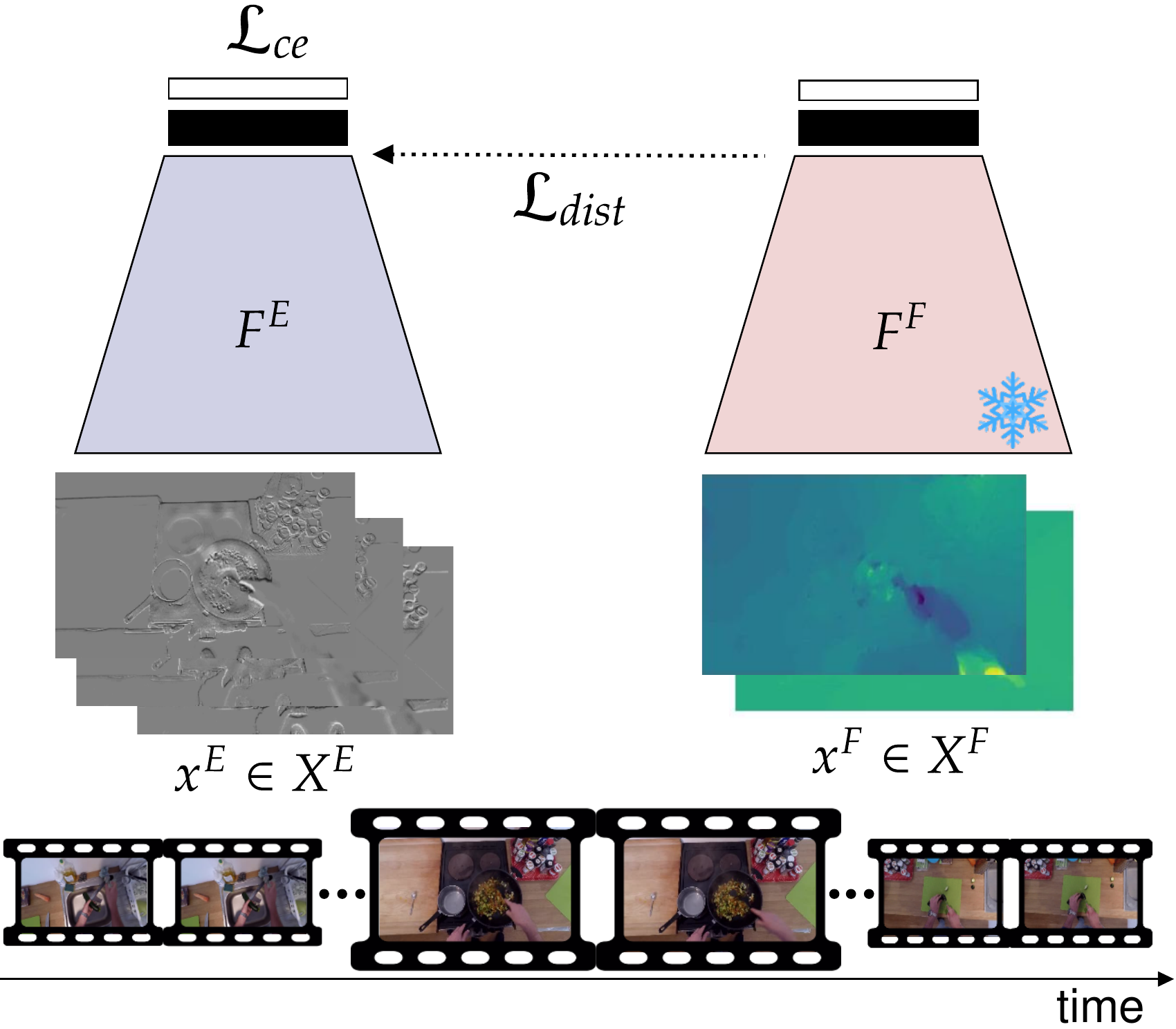}
    \vspace{-0.2cm}
    \caption{Illustration of the proposed E$^2$(GO)MO. The input $\mathbf{x}^E$ and $\mathbf{x}^F$ from the event and flow modality are passed to the feature extractors $F^E$ and $F^F$ respectively. Information from the pre-trained teacher stream (frozen) $F^F$ is distilled to the student stream $F^E$. The latter is trained with standard cross-entropy loss. }
    \vspace{-0.4cm}
    \label{fig:my_label}
\end{figure}




While a traditional RGB frame encodes static information only, frame-based representations used for event data also carry motion information on the channel dimension (see Section \ref{ChallengesEventData}). Indeed, each temporal channel encodes the motion that occurs in the blind-time between a pair of standard frames of the video recording. We propose two different approaches to make standard CNNs able to exploit this information. The first, which we name E$^2$(GO), explicitly models temporal relationships by introducing channel operations that promote motion reasoning. The second, instead, uses a student-teacher strategy that we call E$^2$(GO)MO to encourage the network to extract motion features during training by utilizing a pre-trained optical flow based network. We detail the two approaches in the following.



\subsection{E$^2$(GO): Event Motion}

In order to enable standard CNNs to capture motion information from event data, we propose two simple but effective architectural variations, which improve the capability of extracting temporal inter-channel relations in 2D and 3D CNNs. We refer to them as E$^2$(GO)-2D and E$^2$(GO)-3D, respectively. 

\paragraph{E$^2$(GO)-2D. }
A common practice in the literature is to extract temporal correlations at video level by modeling dependencies between different frames \cite{kazakos2019epic,lin2019tsm}. A peculiarity of event representation is that the channel sequence encodes continuous motion, thus describing micro-movements in the scene. This observation motivates us to extend the practice of modeling temporal relations to also learn short-range correlations between event channels.


We propose to do this by exploiting \textit{Squeeze And Excitation} modules~\cite{hu2018squeeze} to enhance attention correlations between channels in 2D CNNs. 
Given an event volume $\mathbf{x}^E \in \mathbb{R}^{T \times H \times W \times F}$ as input,  where $T$ is the temporal dimension, $H \times W$ is the feature map resolution and $F$ indicates the number of channels, we refer as ${\mathbf{f}_i}^E \in \mathbb{R}^{T \times H_i \times W_i \times C_i} $ to the features extracted from the $i$-th layer of the network. As a first step, we ``squeeze" the spatial information content of $\mathbf{f}_i^E$ into a channel descriptor by performing feature aggregation along the spatial dimensions. 
It follows an ``excitation" operator, which takes in input $\mathbf{z}^E_{sq}$ to produce an activation vector $\mathbf{s}$ to be used to scale $\mathbf{x}^E$. The scaling vector $\mathbf{s}$ is obtained from $\mathbf{z}_{sq}^E$ through two fully-connected layers with a bottleneck that down sizes $C$ to $C/r$. Finally, $\mathbf{s}$ is used to re-weight $\mathbf{x}^E$, resulting in a new feature vector $\mathbf{\tilde{x}}^E$ to enhance discriminative motion features and discard the less informative ones.  
As a result, $\mathbf{\tilde{x}}^E$ encodes the relation dynamics between different temporal channels, effectively modeling the dependencies between them as a result of a self-attention function on channel dimension.  

\paragraph{E$^2$(GO)-3D. }
Similarly, we propose to exploit 3D-CNNs' ability to process temporal information through a 3D kernel. Starting from the same input $\mathbf{x}^E \in \mathbb{R}^{T \times H \times W \times F}$, traditional 3D CNNs apply a 3D convolution on the $(T,H,W,F)$ dimensions, resulting in an output of shape $(T',H',W',C)$. We re-purpose the 3D convolution operator in this context to operate on $\mathbf{x}^E \in \mathbb{R}^{(F\cdot T) \times H \times W \times 1}$ by moving the channel dimensions on the temporal axis. The convolution directly models the micro-movements contained across the temporal channels of the event representation, which would otherwise be ignored when processed on the channel dimension.

\subsection{E$^2$(GO)MO: Learning from Flow}
\label{LearningFromFlow}
Our goal is to train a network using both event and optical flow data, avoiding the need to estimate the latter during testing. Given a multi-modal input $X=(X^E,X^F)$, where $X^E$ denotes the event modality and $X^F$ denotes the flow one, we indicate with $F^E$ and $F^F$ their respective feature extractors, and the resulting features with $\mathbf{f}^E=F^E(\mathbf{x}^E)$ and $\mathbf{f}^F=F^F(\mathbf{x}^F)$. As a first step, we train the flow extractor $F^F$ using a cross-entropy loss between the true action labels $\hat{y}$ and the labels $y^F$ predicted by a fully connected layer on top of $F^F$. 
Then, we first freeze the flow stream $F^F$, and then train the event stream $F^E$ by combining the standard cross-entropy loss with a \textit{distillation loss} defined as the $L_2$ between features $\mathbf{f}^E$ and $\mathbf{f}^F$: 
%
\begin{equation}
    \mathcal{L}_{dist}= \alpha || \mathbf{f}^E- \mathbf{f}^F || ^2 .
\end{equation}
where $\alpha$ is a scaling hyperparameter.
Such loss encourages features of the event stream to match those of the flow one, forcing $F^E$ to mimic the behavior of $F^F$, and thus enabling the two to produce similar activations. Notice that we use optical flow data only during training and remove the teacher branch during inference, thus exploiting the advantages of this modality but effectively avoiding its computational complexity in prediction.

\section{Experiments}
In this section, we first introduce the experimental setup used (Section \ref{sec:6.1}), then we benchmark event data and validate the proposed E$^2$(GO) and E$^2$(GO)MO. We conclude the section with a discussion and limitation paragraph. 
\subsection{Experimental Setup}\label{sec:6.1}
 
 \paragraph{Input.} Experiments with I3D~\cite{carreira2017quo} are conducted by sampling one random clip from the video during training and 5 equidistant clips spanning across all the video during test, as in~\cite{munro2020multi}. The number of frames composing each clip is 16 for RGB and optical flow, and 10 for events. 
 For TSN~\cite{wang2018temporal} and TSM~\cite{lin2019tsm} architectures, uniform sampling is used, consisting of 5 frames uniformly sampled along the video. During testing, 5 clips per video are adopted, following
 ~\cite{lin2019tsm}. The Voxel Grid representations are clipped between $-0.5$ and $0.5$, and all data modalities are rescaled and normalized in accordance with the pretrained network associated with the architecture adopted. For all modalities, we used standard data augmentation following~\cite{wang2016temporal}.

\paragraph{Implementation and Training Details.}
With regard to I3D, the original implementation from~\cite{carreira2017quo} has been chosen, while TSN and TSM models have been built using respectively a BN-Inception~\cite{ioffe2015batch} and a ResNet-50~\cite{he2016deep} backbone. 
In the multi-modal experiments, a classic late fusion strategy is used, in which prediction scores from different modalities are summed and the error is backpropagated to all modalities.
All models are implemented in PyTorch~\cite{pytorch}.
SGD with momentum~\cite{sgdmom} with a starting learning rate $\eta$ of $0.01$, a weight decay of $10^{-7}$ and a momentum $\mu$ of $0.9$ is used as optimizer. We trained the networks for a total of $5000$ iterations with a learning rate decay to $1e{-3}$ at step $3000$. All the experiments are performed with a batch size of 128 on 4 NVIDIA Tesla V100 16Gb GPUs. For the distillation loss, we found the best hyperparameter $\alpha=100$ (see Supplementary for details). As far as the evaluation protocol used, for \textit{seen} results we train on kitchen $D_i$ and test on the same ($D_i \rightarrow D_i$), $i \in \{1,2,3\}$. We evaluate performance on \textit{unseen} test by training on $D_i$ and testing on  $D_j$, with $i \neq j$ and $i, j \in \{1,2,3\}$ ($D_i \rightarrow D_j$). 

\subsection{Results}

\paragraph{Event Analysis.} In Table \ref{tab:channels} we show the performance of events on the three selected 
action recognition architectures (see Section \ref{ChallengesEventData}). 
We observed that extracting 3-channels Voxel Grid is the optimal choice and we used it in all the remaining experiments (more details in Supplementary). 
Considering the performance on both seen and unseen test sets, the TSM model is the one performing the best, while I3D performs slightly worse. One explanation is that it only processes a small portion of the video at a time, catching only local features when trained at the clip level. 
TSM, on the other hand, can capture global features because it works with frames that cover the full video. The poor performance of TSN is to be expected, given that its frame aggregation prevents any temporal correlation from being modeled.
Thus, unless otherwise stated, we perform video-level analysis and evaluate the proposed approaches on TSM and I3D backbones in all of the following experiments.


\begin{table}[t]
\begin{minipage}{\linewidth}
\centering
\begin{adjustbox}{width=0.9\columnwidth, margin=0ex 1ex 0ex 0ex}

\begin{tabular}{cclcc}
\toprule\noalign{\smallskip}
\multicolumn{1}{l}{\textbf{Model}} & \multicolumn{1}{l}{\textbf{Voxel ch.}} & \textbf{Testing} & \multicolumn{1}{l}{\textbf{Seen (\%)}} & \multicolumn{1}{l}{\textbf{Unseen (\%) }} \\ \hline
\multirow{2}{*}{I3D}      & \multirow{2}{*}{3}                & Clip    & 53.75                          & 35.90                         \\  
                           &                                   & Video   & \textbf{55.54}                 & \textbf{37.52}                \\
                           \hline
\multirow{2}{*}{TSN}       
                           & \multirow{2}{*}{3}                & Clip    & 58.81 &	34.65                         \\
                           &                                   & Video   & \textbf{59.82} &	\textbf{35.24}                 \\ \hline
\multirow{2}{*}{TSM}       
                           & \multirow{2}{*}{3}                & Clip    & 64.38               & 37.75               \\
                           &                                   & Video   & \textbf{65.93}                         & \textbf{38.23}                         \\ 
                           \bottomrule

\end{tabular}

\vspace{-0.5cm}

\end{adjustbox}
\caption{\textit{Mean} accuracy ($\%$) over all $D_i \rightarrow D_j$ combinations on I3D, TSN and TSM on both seen and unseen test sets.  }
\vspace{-0.5cm}

\label{tab:channels}

\end{minipage}
\end{table}
\begin{table*}[]
\begin{minipage}{\linewidth}
\begin{adjustbox}{width=1\columnwidth, margin=0ex 1ex 0ex 0ex}

\begin{tabular}{llccc|cccccc|cc}

\toprule\noalign{\smallskip}
\textbf{Modality}  & \textbf{Model}   & \textbf{D1} & \textbf{D2} & \textbf{D3} & \textbf{D1$\rightarrow$ D2} & \textbf{D1$\rightarrow$D3} & \textbf{D2$\rightarrow$D1} & \textbf{D2$\rightarrow$D3} & \textbf{D3$\rightarrow$D1} & \textbf{D3$\rightarrow$D2 }& \textbf{Seen} (\%)      & \textbf{Unseen} (\%)     \\ \hline \noalign{\smallskip}

RGB       & I3D     & 53.67 & 61.12 & 60.70 & 34.50 & 35.70 & 34.94 & 36.46 & 33.93 & 38.37 & \textbf{58.49} & 35.65          \\ \hline \noalign{\smallskip}
Event     & I3D     & 50.32 & 58.33 & 57.99 & 37.27 & 39.12 & 32.98 & 36.52 & 35.68 & 43.56 & 55.54          & 37.52          \\
Event     & E$^2$(GO)-3D & 50.52 & 62.99 & 60.11 & 38.07 & 38.71 & 35.02 & 38.49 & 36.73 & 45.53 & 57.87          & \textbf{38.76} \\ \hline
 \hline \noalign{\smallskip}
RGB & TSM & 61.61 & 77.08  & 75.75  & 37.39  &	32.49  &	34.28  & 38.99  & 34.43  & 38.25  &	\textbf{71.48}  & 35.97 
\\ \hline \noalign{\smallskip}

Event & TSM  &56.86 & 72.43 & 68.49 & 28.73 & 34.00 & 37.09 & 42.30 & 42.27 & 45.02 & 65.93 & 38.23         \\
Event & E$^2$(GO)-2D &56.58 & 70.03 & 69.60 & 34.98 & 35.16 & 38.21 & 47.80 & 41.71 & 44.13 & 65.40 & \textbf{40.33} \\
\bottomrule
\end{tabular}

\end{adjustbox}
\end{minipage}
\caption{Accuracy ($\%$) of event w.r.t. RGB on both I3D and TSM. Results are shown on all shifts, i.e., $D_i \rightarrow D_j$ indicates we trained on $D_i$ and tested on $D_j$, and $D_i$ means we trained and test on the same. E$^2$(GO)-3D and E$^2$(GO)-2D improvements are shown w.r.t. to their respective baselines, where no architectural variations are performed. In \textbf{bold} the best results on both seen and unseen for each backbone.  }\label{tab:shift}

\end{table*}

\begin{table}[t]
\begin{adjustbox}{width=\columnwidth, margin=0ex 1ex 0ex 0ex}
\begin{tabular}{lllcc}
\toprule\noalign{\smallskip}
\textbf{Model}          & \textbf{Streams}        & \textbf{Pretrain}        & \multicolumn{1}{l}{\textbf{Seen (\%)}} & \multicolumn{1}{l}{\textbf{Unseen (\%)}} \\ \hline\noalign{\smallskip}
I3D            & Event          & Kinetics-400 (R)  & 55.54                              & 37.52                                \\\noalign{\smallskip}
E$^2$(GO)-3D & Event          & Kinetics-400 (R)  & 57.87                              & 38.76                              \\\noalign{\smallskip}
TSM            & Event          & ImageNet        & \textbf{65.93}  &  38.23                                \\\noalign{\smallskip}

E$^2$(GO)-2D & Event          & ImageNet        & 65.40  & \textbf{40.33}                      \\ \hline\noalign{\smallskip}
I3D            & Event+RGB      & Kinetics-400 (R)  & 59.12                              & 38.13                                \\\noalign{\smallskip}

E$^2$(GO)-3D & Event+RGB      & Kinetics-400 (R)  & 61.23                              & \textbf{41.85}                       \\\noalign{\smallskip}
TSM            & Event+RGB      & ImageNet        & 71.88                              & 39.92                                \\\noalign{\smallskip}

E$^2$(GO)-2D & Event+RGB      & ImageNet        & \textbf{72.42}	 & 40.61                               \\ \hline\noalign{\smallskip}
I3D            & Event+Flow     & Kinetics-400 (R)  & 60.48                              & 44.47                            \\\noalign{\smallskip}

E$^2$(GO)-3D & Event+Flow     & Kinetics-400 (R)  &                             62.66              &          45.86                         \\\noalign{\smallskip}
TSM            & Event+Flow     & ImageNet        &                              72.26 &	46.89                                           \\\noalign{\smallskip}

E$^2$(GO)-2D & Event+Flow     & ImageNet        & \textbf{72.87} &	\textbf{49.23}                              \\\hline \noalign{\smallskip}
I3D            & RGB+Flow     & Kinetics-400 (R)  & 62.07 &	44.56          \\\noalign{\smallskip}
TSM            & RGB+Flow     & ImageNet        &                          \textbf{75.08}	 & \textbf{	45.66}                                           \\

\bottomrule
\end{tabular}
\end{adjustbox}
\caption{Accuracy results ($\%$) of the event modality when used in combination to stardard RGB and optical flow. In \textbf{bold} the best result for each modality combination. }
\label{tab:event}
    \vspace{-0.2cm}

\end{table}

\begin{table*}[h]
\begin{minipage}{\linewidth}
\begin{adjustbox}{width=1\columnwidth, margin=0ex 1ex 0ex 0ex}
\begin{tabular}{lc|rrr|rrrrrr|cc|l}
\toprule\noalign{\smallskip}

         Method & Model  & \multicolumn{1}{c}{\textbf{D1}} & \multicolumn{1}{c}{\textbf{D2}} & \multicolumn{1}{c|}{\textbf{D3}} & \multicolumn{1}{c}{\textbf{D1$\rightarrow$D2}} & \multicolumn{1}{c}{\textbf{D1$\rightarrow$D3}} & \multicolumn{1}{c}{\textbf{D2$\rightarrow$D1}} & \multicolumn{1}{c}{\textbf{D2$\rightarrow$D3}} & \multicolumn{1}{c}{\textbf{D3$\rightarrow$D1}} & \multicolumn{1}{c|}{\textbf{D3$\rightarrow$D2}} & \multicolumn{1}{l}{\textbf{Seen (\%)}} & \multicolumn{1}{l|}{\textbf{Unseen (\%)}} & \multicolumn{1}{l}{\textbf{Mean (\%)}} \\ \cline{0-13} \noalign{\smallskip}

RGB & TSM  & 61.61 & 77.08 & 75.75 & 37.39 & 32.49 & 34.28 & 38.99 & 34.43 & 38.25 & 71.48 & 35.97 & 53.73  \\ \noalign{\smallskip}
RGB + $ \mathcal{L}_{dist}$ & TSM    & 63.36 & 79.47 & 77.97  & 38.61 & 35.73 & 39.36 & 41.09  & 34.76  & 49.68 & \textbf{73.60} & {39.87} & {56.73} \mytriangle{Green}\textcolor{Green}{$\mathbf{+3}$}  \\ \noalign{\smallskip}  \hline \noalign{\smallskip}

\rowcolor{Gray} RGB + Flow & TSM & 66.97 & 79.69 & 78.58 & 43.76 & 43.76 & 45.80 & 47.13 & 45.44 & 48.09 & \underline{75.08} & 45.66 & 60.37
   \\ \noalign{\smallskip} \hline\hline \noalign{\smallskip}
Event & TSM    & 56.86 & 72.43 & 68.49 & 28.73 & 34.00 & 37.09 & 42.30 & 42.27 & 45.02 & 65.93 & 38.23 & 52.08 \\ \noalign{\smallskip}

Event & E$^2$(GO)-2D    & 56.58 & 70.03 & 69.60 & 34.98 & 35.16 & 38.21 & 47.80 & 41.71 & 44.13 & 65.40 & 40.33 & 52.87 \\ \noalign{\smallskip}
Event & E$^2$(GO)MO-2D   &61.38 & 75.83 & 75.08 & 39.77 & 37.19 & 44.71 & 51.03 & 47.01 & 53.73 & {70.76} & \textbf{45.57} & \textbf{58.17} \mytriangle{Green}\textcolor{Green}{$\mathbf{+5.3}$} \\\hline \noalign{\smallskip}

\rowcolor{Gray} Event + Flow & E$^2$(GO)-2D  & 65.11 & 77.58 & 75.91 & 42.12 & 41.80 &	48.20 & 53.50 & 51.85  & 57.91  & 72.87 & \underline{49.23} & \underline{61.05}

  \\

    \bottomrule

\end{tabular}

\end{adjustbox}
\end{minipage}
\vspace{-0.15cm}
\caption{Accuracy (\%) of E$^2$(GO)MO w.r.t. the baseline on events (TSM) and E$^2$(GO)-2D. We compare E$^2$(GO)MO with the same approach on RGB to validate the choice of combining event and flow. In \textbf{bold} the best uni-modal, \underline{underlined} the best multi-modal. }\label{tab:KD}
\vspace{-0.2cm}
\end{table*}

\vspace{-0.2cm}

\paragraph{Event \textit{vs} RGB.} 
In Table \ref{tab:shift} we compare events against the RGB modality.
Results show that events surpass RGBs by up to $3\%$ on unseen test sets. 
Indeed, it has been shown in the literature that appearance-based CNNs are biased toward texture, which causes them to underperform across-domain, but their robustness improves when shape-bias is increased~\cite{geirhos2018imagenet}. 
We believe this is the primary reason why event representations, focusing more on geometric and temporal information rather than texture variations, are more invariant to domain changes.
The same considerations also apply to seen tests, where RGB-based networks overfit by leveraging domain-specific features.
We remark that until now the event modality was still lagging behind RGB images in purely visual tasks, as reported by the recent release of N-ImageNet benchmark~\cite{kim2021n}, where the best performing event architecture scores $48.94\%$, considerably below RGB's $>90\%$ accuracy~\cite{he2021masked,dai2021coatnet,zhai2021scaling,pham2021meta}.
In this study, instead, we show that events can outperform RGBs in challenging unseen scenarios and compete under seen ones, emphasizing their importance in egocentric vision.

\paragraph{E$^2$(GO).} 

In Table \ref{tab:shift} we show the performance of E$^2$(GO)-2D and E$^2$(GO)-3D. Those are beneficial especially on unseen test sets, as they aim to enhance temporal correlations, thus allowing the network to emphasise motion features that are informative while suppressing those that are not correlated with the action. E$^2$(GO)-3D achieves an improvement by up to $2\%$ on seen test set, while E$^2$(GO)-2D achieves results on-par with the baseline TSM. This could be explained by the fact that 2D CNNs, being based on frame-based techniques, rely heavily on visual signals. Indeed, while those are harmful when changing environments, they can be helpful on seen ones. I3D, on the other hand, is naturally more responsive to temporal correlations. 
Extending its temporal reasoning to micro-movements facilitates it in extracting discriminative features for the action, reflecting in an higher accuracy even when testing on the same environment.

\paragraph{Multi-Modal Analysis.} 
Table \ref{tab:event} illustrates the behavior of the event modality when combined with RGB and optical flow. 
When combined with RGB, it achieves an improvement of up to $7\%$ on seen test sets and $3\%$ on unseen ones. When combing events with optical flow, the best performance is achieved, improving event results by up to $7\%$ on seen domains and $9\%$ on unseen ones. This suggests that, while both event and flow encode motion, flow emphasizes the motion-relevant part, neglecting the scene or object affordances, while the event data maintain useful information about objects' shape (see Figure \ref{fig:kitchens}).
For this reason, the event modality is potentially more convenient to be combined with optical flow data than with RGB, which, instead, suffer on unseen domains due to its dependency on appearance. It is also worth noting that it outperforms standard RGB+Flow since standard event representations does not emphasize features of appearance as much as RGB does. 






\paragraph{E$^2$(GO)MO.} 
In Table \ref{tab:KD} we illustrate the performance of E$^2$(GO)MO against an RGB-based TSM, which we proved to be the most robust architecture in the previous analysis.
To prove our claim that the proposed {distillation} technique benefits from motion features, we also apply the same mechanism to an RGB-based stream, which we label in Table \ref{tab:KD} with the RGB+$ \mathcal{L}_{dist}$ entry.
Both event and RGB benefit from the flow learning strategy, improving performance on unseen tests (+5.3\% and +3\% respectively), confirming the importance of motion information in real-world scenarios. 
However, E$^2$(GO)MO  gains far more from the distillation loss $\mathcal{L}_{dist}$ than RGB, indicating that event data conveys more motion-rich features than RGB streams, thus proving our argument.
Finally, we compare these two networks against their multi-modal upper bound performance, obtained exploiting the offline-computed optical flow also in prediction, namely RGB+Flow and E$^2$(GO)+Flow.
Despite both are unable to reach their upper bound, E$^2$(GO)MO is much closer to E$^2$(GO)+Flow, and it even exceeds the multi-modal RGB-Flow performance. This result further motivates the use of event data 
in egocentric vision.

   \begin{figure}[t]
    \centering
    \includegraphics[width=\columnwidth]{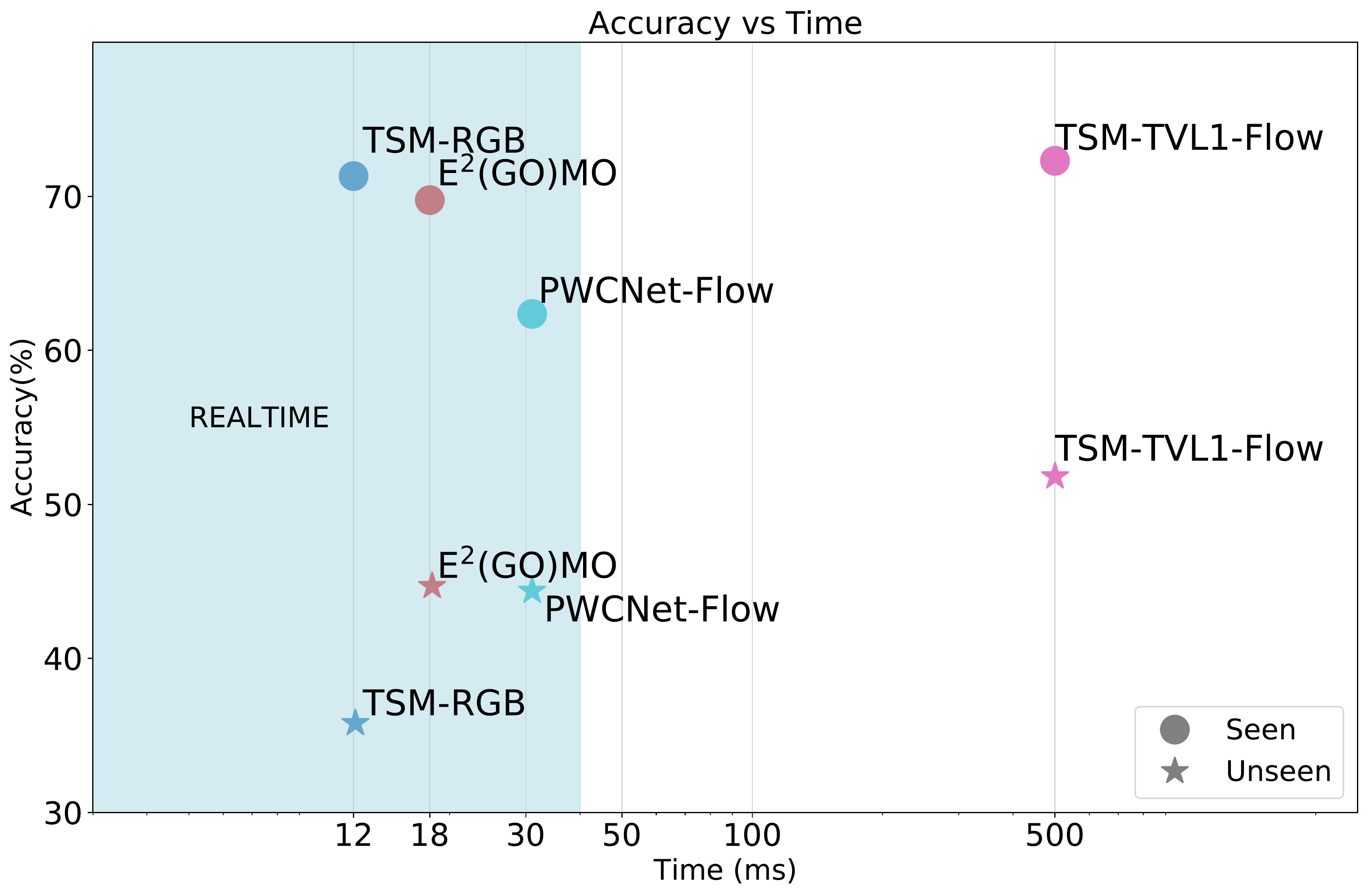}
    \vspace{-0.5cm}
    \caption{Accuracy \textit{vs} time of RGB modality, E$^2$(GO)MO, estimated PWCNet optical flow and TV-L1 optical Flow on seen and unseen scenarios for one clip evalutation. }
        \vspace{-0.2cm}

    \label{fig:time}
\end{figure}
 \begin{table}[]
\begin{minipage}{\linewidth}
\centering
\begin{adjustbox}{width=0.85\columnwidth, margin=0ex 1ex 0ex 0ex}

\begin{tabular}{lccccc}
\toprule\noalign{\smallskip}

\textbf{Stream}         & \textbf{Model}  & \textbf{Repr.} & \textbf{Seen} & \textbf{Unseen} \\
&  & \textbf{Time (ms)} & \textbf{(\%)} & \textbf{(\%)} \\
\hline\noalign{\smallskip}

RGB & I3D  & &  \textbf{58.49} & 35.65  \\
Event          & I3D   & \textbf{6ms}        &                               55.54 &          	   37.52 \\
Event          & E$^2$(GO)-3D   & \textbf{6ms}        &                               57.87 &          	    \textbf{38.76} \\ 
 
\rowcolor{Gray}
Flow (TV-L1)   & I3D     & \textcolor{red}{488ms}     &                        58.47& 	43.40                           \\
\noalign{\smallskip}\hline \noalign{\smallskip}
RGB & TSM && \textbf{71.48} & 35.97  \\ 
Event          & TSM    & \textbf{6ms}        & 65.93	                        & 38.23   \\ 
Event          & E$^2$(GO)-2D   & \textbf{6ms}        & 65.40                      & 	\textbf{40.33}   \\ 
\rowcolor{Gray}
Flow (TV-L1)   & TSM    & \textcolor{red}{488ms}     & 73.23	& 53.98    \\
\bottomrule
\end{tabular}
\end{adjustbox}
\end{minipage}
\vspace{-0.3cm}
\caption{Accuracy result of RGB, Event and optical flow (TV-L1), along with their representation time, i.e., time to calculate the Voxel Grid for event, and extraction time for TV-L1 flow.  }\label{tab:EF}

\vspace{-0.2cm}

\end{table}

\paragraph{Event \textit{vs}. Optical Flow.} We illustrate in Figure \ref{fig:time} the accuracy \textit{vs}. average time per frame trade-off at test time on both seen and unseen data. We report performance with both TV-L1 flow, computed offline~\cite{zach2007duality}, and the one extracted from PWC-Net~\cite{sun2018pwc}. The latter is the most competitive among existing end-to-end CNN models for flow, providing an optimal balance between time and accuracy. 
For calculation, we use a NVIDIA Titan RTX GPU, and report both input's computation and forward time, ignoring data access time.
We also highlight the range under which we can perform real-time action recognition, using the threshold considered in~\cite{song2016fast} to determine a sufficient frame (sampling) rate for a motion tracking system as a reference point. The plot clearly shows how TV-L1 achieves higher accuracy at the cost of 488 ms of extraction time, making it unsuitable for online scenarios. 
When the optical flow is estimated online with PWC-Net, performance drops dramatically (by up to $10\%$ on seen tests and $8\%$ on unseen tests). 
Additionally, PWC-Net necessitates the execution of an additional network, increasing the parameter count ($\approx$ 40M) and requiring an additional fine-tuning stage. 
In contrast, we do not have to compute flow at test time, thus we can take full advantage of the more precise optical flow when distilling. 
Despite E$^2$(GO)MO does not explicitly use flow during inference, it still outperforms PWC-Net on seen tests (by up to $6\% $) and performs on par with it on unseen ones.

 \paragraph{Discussion and Limitations.}

 
As it is currently not possible to fully replicate event camera behaviors, event simulation may create undesirable sim-to-real domain shift~\cite{planamente2021da4event,stoffregen2020reducing}. 
Nevertheless, several works showed that simulated events are robust enough to generalize well to real ones~\cite{planamente2021da4event,stoffregen2020reducing,gehrig2020video}. 
As we introduce event data in egocentric action recognition for the first time, we aim at providing a direct comparison with common benchmarks in the literature~\cite{fathi2012learning,damen2018scaling,damen2021rescaling} and place the event modality in a competitive setting against well-established modalities. These aspects motive us to simulate the event data instead of generating a new first-person dataset from scratch.

Starting from the promising results of our work, we plan to further explore the use of real event streams in this context in order to validate the considerations done so far on a real camera. 
Moreover, Table \ref{tab:EF} shows that, despite its high computational and time cost, TV-L1 optical flow still demonstrates higher performance, especially an extraordinary resiliency to domain changes. We primarily attribute this to the fact that the algorithm for extracting it partially filters out camera motion, resulting in cleaner motion data compared to the unprocessed events. To this purpose, interesting future works could involve the exploitation of motion compensation techniques commonly used with events~\cite{stoffregen2019event,stoffregen2019event} to remove redundant background noise.

\section{Conclusion}
In this paper, we presented N-EPIC-Kitchens, the first event-based egocentric action recognition dataset. Exploiting the variety of data modes at our disposal, we carried out an in-depth comparative analysis whose results demonstrate the relevance of motion information in action recognition context. 
Given these findings, we proposed and evaluated two novel approaches suited for event data (E$^2$(GO) and E$^2$(GO)MO) that, by emphasizing motion information, produced competitive results compared to the computational expensive optical flow modality. Through extensive experiments, we bring to light the robustness of event data and their applicability in an online action recognition setting, pushing the community to further explore in this direction. 

{\small 
\textbf{Acknowledgements.}
This work was supported by the CINI Consortium through the VIDESEC project.
We acknowledge the CINECA award under the ISCRA initiative and IIT HPC infrastructure, for the availability of high performance computing resources and support. A special thanks goes to Antonio Loquercio for his precious advice.}



{\small
\bibliographystyle{ieee_fullname}
\bibliography{egbib}
}

\end{document}